%% file: root.tex
\documentclass[letterpaper, 10 pt, conference]{ieeeconf} 

\IEEEoverridecommandlockouts 
\newif\ifanonymous
\anonymousfalse 

\usepackage{amsmath,amssymb}
\usepackage{graphicx}
\graphicspath{{assets/}}

\usepackage{booktabs}
\usepackage{multirow}
\usepackage{xspace}
\usepackage[table,dvipsnames]{xcolor}
\usepackage{url}
\usepackage{cite}
\usepackage{censor} 

\usepackage{soul}
\sethlcolor{yellow}

\usepackage{algorithm}
\usepackage{algorithmic}
\usepackage{pifont}
\newcommand{\xmark}{\ding{55}}%

\usepackage[caption=false,font=footnotesize]{subfig}
\usepackage{capt-of} 

\usepackage[
    colorlinks=true,
    linkcolor=Orange,
    citecolor=blue,
    urlcolor=Orange]{hyperref}
\usepackage[capitalise,nameinlink]{cleveref}

\newcommand{\model}{CodeGraphVLP\xspace}

\title{\LARGE \bf
CodeGraphVLP: Code-as-Planner Meets Semantic-Graph State for Non-Markovian Vision-Language-Action Models}

\ifanonymous
\author{Anonymous Author(s)}
\else
\author{Khoa Vo$^{\dagger,1}$, Sieu Tran$^{\dagger,1}$, Taisei Hanyu$^{1}$, Yuki Ikebe$^{1}$, Duy Nguyen$^{2}$, Nghi D. Q. Bui$^{3}$, Minh Vu$^{4}$\\Anthony Gunderman$^{1}$, Chase Rainwater$^{1}$, Anh Nguyen$^{5}$, and Ngan Le$^{1}$%
\thanks{$^{\dagger}$Authors marked with this symbol contributed equally. $^{1}$University of Arkansas, Fayetteville, AR, USA. $^{2}$Max Planck Research School for Intelligent Systems and the University of Stuttgart, Stuttgart, Germany. $^{3}$Center of AI Research, VinUniversity, VietNam. $^{4}$TU Wien, Vienna, Austria. $^{5}$University of Liverpool, Liverpool, U.K.}%
}
\fi

\IEEEaftertitletext{\input{sections/00_teaser}}

\begin{document}

\bstctlcite{BSTcontrol}
\maketitle
\thispagestyle{empty}
\pagestyle{empty}

\begin{abstract}
\input{sections/00_abstract}
\end{abstract}

\section{Introduction}
\input{sections/01_introduction}

\section{Related Work}
\input{sections/02_related_work}

\section{Method}
\input{sections/03_method}

\section{Experiments}
\input{sections/04_experiments}

\section{Conclusion}
\input{sections/05_conclusion}

\ifanonymous
\else
\fi


\bibliographystyle{IEEEtran}
\bibliography{references}

\end{document}

%% file: sections/00_teaser.tex
\begin{center}
    \includegraphics[width=0.98\linewidth]{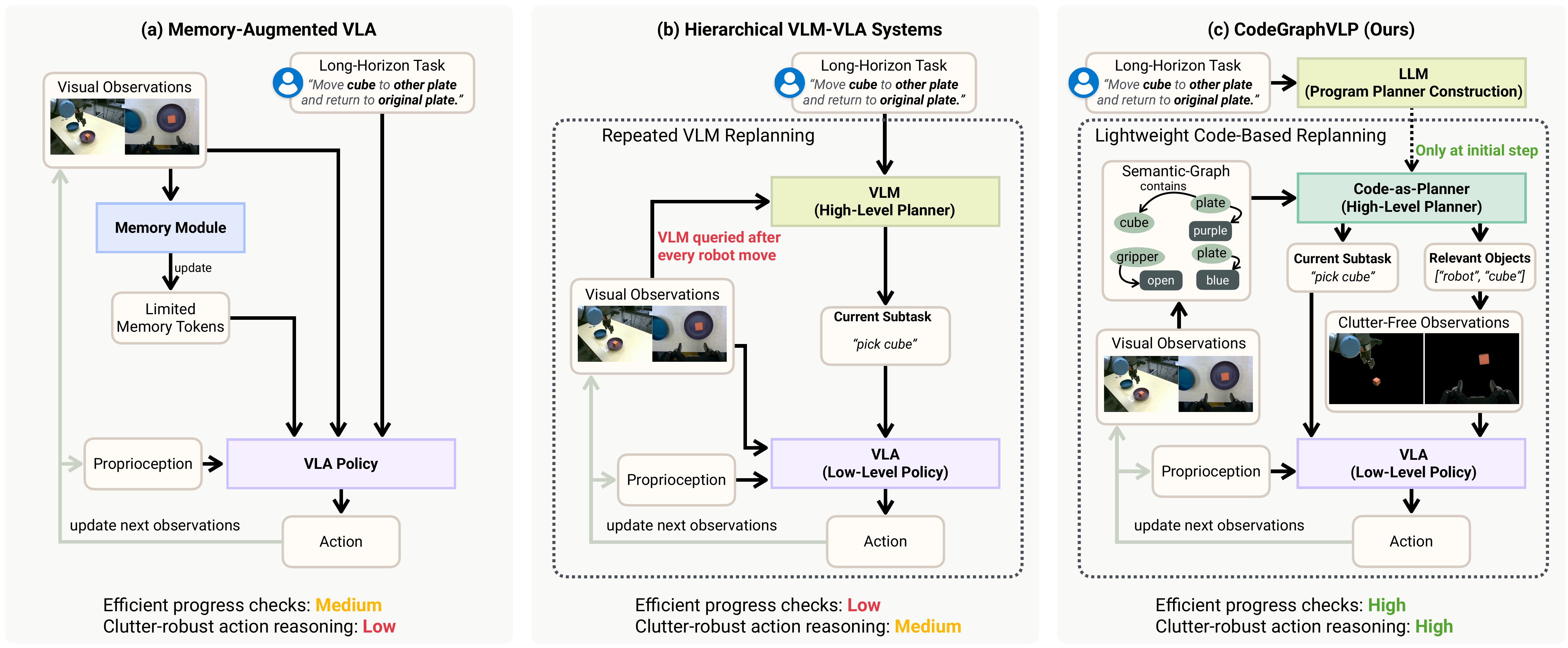}
    \vspace{-0.6em}
    \captionof{figure}{\textbf{Architectures for non-Markovian long-horizon manipulation.} \textbf{(a) Memory-augmented VLA} equips a short-horizon policy with memory context, offering moderately efficient progress checks and limited robustness for action reasoning in clutter. \textbf{(b) Hierarchical VLM--VLA} uses a VLM planner to reason about subtasks and guide a VLA policy with subtask-level cues, improving clutter robustness but incurring highlatency due to repeated VLM inference online.
    \textbf{(c) \model (Ours)} performs fast progress checks by synthesizing Code-as-Planner just once at initial step and executes it
    over a persistent semantic graph to produce clutter-free visual--language prompts to the VLA. Together, this enables efficient progress checks and reliable action reasoning in cluttered scenes.
    }
    \label{fig:teaser}
\end{center}

%% file: sections/00_abstract.tex
Vision-Language-Action (VLA) models promise generalist robot manipulation, but are typically trained and deployed as short-horizon policies that assume the latest observation is sufficient for action reasoning. This assumption breaks in non-Markovian long-horizon tasks, where task-relevant evidence can be occluded or appear only earlier in the trajectory, and where clutter and distractors make fine-grained visual grounding brittle. We present \model, a hierarchical framework that enables reliable long-horizon manipulation by combining a persistent semantic-graph state with an executable code-based planner and progress-guided visual--language prompting. The semantic-graph maintains task-relevant entities and relations under partial observability. The synthesized planner executes over this semantic-graph to perform efficient progress checks and outputs a subtask instruction together with subtask-relevant objects. We use these outputs to construct clutter-suppressed observations that focus the VLA executor on critical evidence. On real-world non-Markovian tasks, \model improves task completion over strong VLA baselines and history-enabled variants while substantially lowering planning latency compared to VLM-in-the-loop planning. We also conduct extensive ablation studies to confirm the contributions of each component.

%% file: sections/01_introduction.tex
Vision-Language-Action (VLA) models~\cite{rt1_2022, rt2_corl2023, octo_rss2024, roboflamingo_iclr2024, openvla_corl2024, pi0_2024, fast_2025, pi05_2025, gr00t_2025} have rapidly advanced generalist robot control by post-training Vision-Language Models (VLMs) backbone for action prediction on large-scale robotic manipulation data~\cite{bridgedatav2_corl2023, oxe_corl2024, droid_2024}. Most VLAs are trained and deployed as short-horizon policies that map the \emph{latest} sensory input and instruction to the next action chunk, implicitly assuming that the current observation is sufficient for action reasoning. However, this assumption is fragile and limits real-world deployment, where long-horizon embodied tasks are intrinsically \emph{non-Markovian}: the information required to choose the next action may not be contained in the latest observation due to partial observability, occlusions, and evolving task context. In other words, the correct next action can depend on task context and past observations.

Two strategies have emerged to extend VLAs to long-horizon tasks. \textit{Memory-augmented VLAs}~\cite{hdm_icra2022,memer_arxiv2025,hamlet_iclr2026,memoryvla_iclr2026} (\Cref{fig:teaser}(a)) support temporally referential grounding by integrating temporal context via step-wise histories, condensed memory tokens, or retrieval of task-relevant past experience. In parallel, \textit{hierarchical VLM--VLA}~\cite{pi05_2025,hirobot_icml2025,thinkact_neurips2025,fast-thinkact_arxiv2026,mindexplore_iccv2025} (\Cref{fig:teaser}(b)) place a VLM planner above a VLA executor, \textit{repeatedly selecting a subtask} and conditioning the VLA to execute it as a short closed-loop segment. This hierarchy is appealing because it reuses VLAs as generalist controllers while leveraging a general-purpose VLM for long-horizon reasoning and progress estimation.

Despite these advances, key non-Markovian challenges remain. Memory-augmented VLAs trade off capacity against efficiency, since scaling history (or retrieval) increases compute and latency, while bounded memory can miss the sparse past evidence needed for correct action selection. Hierarchical VLM--VLA systems reduce the burden on the executor by letting a VLM carry long-horizon reasoning and progress estimation, but they typically do not maintain a persistent grounded task state. Decisions therefore still rely heavily on the latest observation and language prompts, and the language-only interface can make the VLA brittle to clutter and distractors.
Furthermore, keeping a VLM in the loop also adds substantial inference latency, undermining real-time manipulation feasibility. These limitations motivate a central question: \emph{How to maintain a compact, explicit task state that preserves task-relevant evidence across time and supports reliable long-horizon control, while keeping inference practical for closed-loop robotics?}

To this end, we propose \model (\textbf{CodeGraph} with \textbf{V}isual--\textbf{L}anguage \textbf{P}rompting) (\Cref{fig:teaser}(c)), a hierarchical framework for non-Markovian long-horizon manipulation.
Our approach integrates: (i) an executable \textbf{code-based planner} for efficient long-horizon progress estimation, (ii) a persistent \textbf{semantic graph state} that encodes task-relevant entities, attributes, and relations under partial observability, and (iii) \textbf{progress-guided visual–textual prompting} for attentive VLA execution in cluttered scenes.

\model maintains a compact semantic-graph that persists across time and is updated online from observations to explicitly capture the evolving task context. Inspired by embodied programmatic control~\cite{codeaspolicies_icra2023}, we synthesize a task-specific code planner that queries this graph to check task progress through state-transition constraints and produce the next subtask instruction along with subtask-relevant objects. This programmatic reasoning provides a \textit{lightweight mechanism for reasoning about task configurations} while avoiding repeated VLM calls for progress checks. We use these outputs to form progress-guided visual--textual prompts that suppress clutter and steer the VLA policy toward task-critical evidence for reliable action reasoning.

We evaluate \model on three real-world tabletop manipulation tasks with non-Markovian dependencies, including one task that additionally emphasizes clutter-robust action reasoning.
\model achieves more reliable task completion than strong VLA baselines, memory-augmented variants, and VLM-in-the-loop alternatives, while substantially reducing planning latency compared to VLM-in-the-loop planning (\Cref{sec:exp:ablation}).
Overall, our key contributions are summarized as follows:

\begin{itemize}[\setlength{\labelindent}{0em}\setlength{\labelsep}{0.4em}]
    \item We introduce \model, which combines a code-based planner, a persistent semantic-graph that explicitly captures the scene, and progress-guided visual--language prompting to enable clutter-robust non-Markovian long-horizon manipulation.
    \item We show how executable code over the semantic graph can track progress and select subtasks efficiently while producing subtask-relevant entities for downstream execution.
    \item We design three real-world tabletop tasks with non-Markovian dependencies and clutter, and show that \model improves task success while reducing planning latency compared to other history-enabled variants. Ablation studies further quantify the benefit of each component in the proposed \model.
\end{itemize}

%% file: sections/02_related_work.tex
\subsection{Vision--Language--Action Models}
\textbf{Short-horizon VLAs.}
Vision--Language--Action (VLA) models~\cite{rt1_2022, octo_rss2024, roboflamingo_iclr2024, openvla_corl2024, pi0_2024, fast_2025, gr00t_2025} build generalist visuomotor policies by adapting pretrained vision--language backbones~\cite{flamingo_neurips2022, palme_icml2023, palix_cvpr2024, prismatic_icml2024, paligemma_2024} on action prediction with large-scale robot demonstrations.
Architecturally, autoregressive VLAs~\cite{openvla_corl2024, fast_2025} discretize actions into tokens and formulate control as next-token prediction, leveraging sequence modeling advances from language models, while \textit{flow-based VLAs}~\cite{pi0_2024,gr00t_2025,pi05_2025} generate continuous trajectories through noise-to-action transformations (e.g., flow matching~\cite{flowmatching_iclr2023}), enabling smoother and higher-frequency control.
Despite recent progress, most VLAs remain largely reactive, conditioning action generation on the latest observation while focusing primarily on low-level control. As a result, long-horizon task progress and persistent context are often left unmodeled, which becomes a bottleneck in non-Markovian settings where decisions depend on past observations that may be occluded or no longer visible.

\begin{figure*}[ht]
    \centering
    \includegraphics[width=0.9\textwidth]{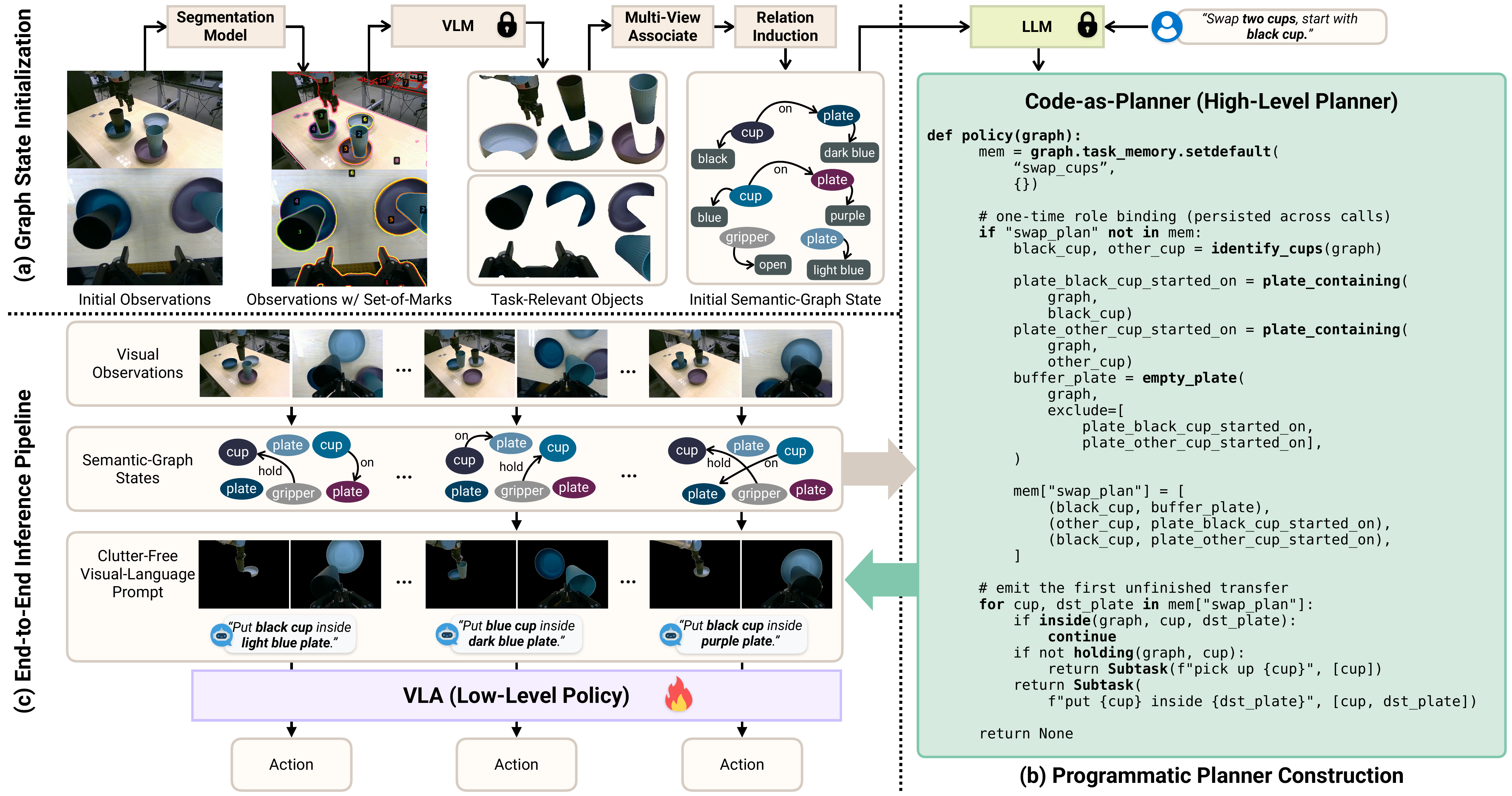}
    \vspace{-0.6em}
    \caption{Overview of \model. \textbf{(a) Graph state initialization}: from the first multi-view observation, we segment objects, identify task-relevant entities, associate them across views, and induce relations to build a compact semantic graph. \textbf{(b) Programmatic planner construction}: given the task and a graph API, an LLM synthesizes an executable \texttt{policy(graph)} that performs progress checks over the graph and maintains lightweight task memory. \textbf{(c) End-to-end inference}: as new observations arrive, we update the semantic graph, execute the planner to produce the next subtask and relevant objects, construct clutter-free visual--language cues, and run a VLA policy for action prediction.
    }
    \label{fig:architecture}
    \vspace{-0.15in}
\end{figure*}

To relax the short-horizon assumption, memory-augmented VLAs incorporate temporal context for history-dependent grounding, using mechanisms such as step-wise memories for temporally referential instructions~\cite{hdm_icra2022}, learned memory tokens~\cite{hamlet_iclr2026,memoryvla_iclr2026}, or retrieval from external experience buffers~\cite{memer_arxiv2025}. However, these approaches typically maintain context as unstructured histories, bounded tokens, or retrieved snippets, limiting their ability to explicitly track long-horizon task progress.

\emph{In contrast, our method externalizes task context into a persistent semantic graph and an code-based planner that queries this structured state to track non-Markovian progress and generate grounded subtask instructions with relevant objects.}

\subsection{Non-Markovian Long-Horizon Reasoning}

\textbf{VLM-based Reasoning.}
A common approach to non-Markovian manipulation uses a pretrained VLM as a high-level reasoning module that converts visual observations and task descriptions into a global plan executed by a downstream controller. Methods in this category explore different prompting and feedback strategies to guide VLM reasoning, including visual prompt signals to highlight task-relevant regions~\cite{moka_rss2024}, feedback from auxiliary modules such as scene description or success detection~\cite{inner_monologue_corl2022}, and visual chain-of-thought reasoning that produces intermediate plans or rationales before action generation~\cite{ecot_2024}. While effective for generating high-level task guidance, these approaches typically reason from a snapshot of the scene and may struggle to adapt when the environment changes or task progress becomes partially observable.

Another line of work performs \textit{closed-loop VLM reasoning}, repeatedly querying the model after executing action chunks to update the plan from new observations. These approaches leverage structured representations and interactive feedback to enable iterative replanning under partial observability~\cite{sayplan_corl2023, pi05_2025,hirobot_icml2025}. While this improves robustness in cluttered and dynamic environments, frequent VLM/LLM queries introduce significant computational overhead. Moreover, the planner typically communicates with the controller through language-only subtask instructions, leaving fine-grained visual grounding to the VLA, which can lead to mis-grounded actions in cluttered scenes.

We overcome such barriers by maintaining a persistent semantic graph state that is updated online and queried by an code-based planner to estimate task progress and generate subtasks, \textit{without repeatedly invoking a VLM}. The planner also outputs \textit{subtask-relevant objects}, enabling clutter-suppressing visual cues that guide the downstream VLA toward grounded action execution.

\textbf{LLM-generated code for policy control.} 
Recent work uses LLMs to synthesize executable programs that interface with perception and robot skills. ~\cite{codeaspolicies_icra2023} generates programs that compose APIs and control primitives to produce reactive policies from language instructions, while ~\cite{rekep_corl2025} generates constraint-based code that solves keypoint-level motion optimization over 3D observations. Rather than generating code to directly control actions or solve motion constraints, we formulate LLM-generated \textit{code as a task-level planner operating over a persistent semantic-graph state} that is updated online. This structured and queryable state enables explicit reasoning about task progress and produces grounded subtasks, allowing the downstream VLA to focus on task-relevant entities and remain robust in cluttered scenes.

%% file: sections/03_method.tex
\subsection{Overview of \model}
\label{sec:problem_statement}

\textbf{Problem formulation.}
Formally, a VLA model is formulated under a Markovian assumption.
At each timestep \(t\), it conditions action generation on the latest observation and the task instruction, without any explicit mechanism to maintain a history-conditioned state.
The robot observes a multi-view RGB observation \(o_t=[I_t^1,\dots,I_t^n]\) together with proprioception \(s_t\) and receives a natural-language instruction \(l\).
Typically, a VLA uses a vision-language backbone to fuse \((o_t,l)\) into a latent representation, followed by an action head that predicts a sequence of \(k\) future control actions (action chunking)~\cite{act_rss2023,diffpolicy_ijrr2025}:
\begin{equation}
\tau_t=(a_t,\dots,a_{t+k-1}) = \pi_\theta(o_t, s_t, l),
\end{equation}
and re-query the model in closed loop as new observations arrive.

This Markovian assumption becomes unreliable in long-horizon tasks with non-Markovian dependencies, where the correct next action can depend on past evidence that may be occluded or absent from \(o_t\).
We capture this by defining the interaction history \(h_t=(o_{0:t}, s_{0:t}, \tau_{0:t-1})\) and writing the optimal policy as
\begin{equation}
\tau_t = \pi^\star(h_t, l).
\end{equation}
However, directly modeling \(h_t\) with longer observation windows, generic memory modules, or iterative VLM-in-the-loop planning is often impractical. Relevant evidence may be sparse or temporally distant and easily confounded by clutter, while memory and compute costs grow poorly with task horizon.

\textbf{Overview of \model.}
Our goal is to enable efficient long-horizon manipulation with VLAs under non-Markovian and cluttered conditions. We introduce \model, which combines a Code-as-Planner with a persistent semantic-graph state that tracks task-relevant objects and relations over time. The code-based planner queries this structured state to estimate task progress and generate grounded subtasks without repeated VLM calls. We further design clutter-free visual–language prompting that exposes the VLA only to subtask-relevant context, improving visual grounding and reducing distractor-induced errors.
An overview is shown in \Cref{fig:architecture}.

\subsection{Persistent Progress State via Semantic-Graph}
\label{sec:semantic_graph}

At the core of our code planner is a persistent semantic-graph state \(\mathcal{G}_t\) that summarizes task-relevant objects and relations, making scene context explicit and queryable.
By filtering out irrelevant clutter, \(\mathcal{G}_t\) remains compact under partial observability.
\\[6pt]
\textbf{Semantic-graph state.}
We define \(\mathcal{G}_t=(\mathcal{V}_t,\mathcal{E}_t)\), where \(\mathcal{V}_t\) is the set of entity nodes and \(\mathcal{E}_t\) is the set of typed relation edges. Each node \(v_i \in \mathcal{V}_t\) represents an entity with a semantic name, per-view mask/box grounding, and task-relevant attributes. Each edge \(e_{ij}=(v_i, v_j, r)\in\mathcal{E}_t\) connects \(v_i, v_j\in\mathcal{V}_t\) with relation type \(r\) (e.g., \emph{in}, \emph{on}, and~\emph{near}).
\\[6pt]
\textbf{Semantic-graph state initialization.}
Given the first observation \(o_0\) and instruction \(l\), we initialize \(\mathcal{G}_0\) in four steps.
\vspace{-0.05in}
\paragraph{Object segmentation}
For each view \(v\), we use an off-the-shelf YOLOE~\cite{yoloe_iccv2025} instance segmenter to extract a set of masks \(\mathcal{M}^v=\{m_j^v\}_j\). We fine-tuned the segmenter with annotated robot arm masks from 10 videos in our teleoperated demonstrations to reliably segment its masks.
\vspace{-0.1in}
 \paragraph{Identify task-relevant objects and their attributes}
We overlay each segmented object with an outline and a numeric ID, following Set-of-Mark prompting~\cite{setofmark_2023}, and query a VLM with the marked frames and instruction \(l\) to identify which objects are relevant to the task and their corresponding attributes. We then remove all irrelevant objects.\\
\vspace{-0.1in}
\paragraph{Multi-view association}
Due to occlusion, each view may contain only a subset of task-relevant objects. We therefore link masks across views into view-consistent graph nodes. We first match objects by \emph{semantic distance}: CLIP visual features~\cite{clip_icml2021} are extracted from each masked region, and we match cross-view pairs whose cosine distance is below a threshold \(\tau_{\mathrm{vis}}\).
Afterwards, we resolve the remaining cross-view associations using a \emph{geometric distance} based on relative spatial layout. Let \(\mathcal{A}\) denote the set of anchors (nodes already matched by semantic distance), and let \(\mathbf{c}(m^v)=(x_m^v,y_m^v)\in\mathbb{R}^2\) be the center-of-mass of mask \(m^v\) of an unmatched object in view \(v\). We compute distance of \(m^v\) to every anchor \(a\in\mathcal{A}\) in the same view as follows:
\begin{equation}
d(m^v,a^v)=\|\mathbf{c}(m^v)-\mathbf{c}(a^v)\|_2
\end{equation}
forming a distance signature \(\mathbf{d}^v(m)=[d(m^v,a^v)]_{a\in\mathcal{A}}\). We further normalize it by the maximum anchor distance in the view,
\(\tilde{\mathbf{d}}^v(m)=\mathbf{d}^v(m)\big/\max_{a\in\mathcal{A}} d(m^v,a^v)\),
to account for scale differences between views.
Finally, we match the remaining objects across views by comparing their normalized signatures \(\tilde{\mathbf{d}}\) and accepting nearest-neighbor matches under a threshold \(\tau_{\mathrm{geo}}\).\\
\vspace{-0.1in}
\paragraph{Relation induction}
With view-consistent nodes in place, relation edges \(\mathcal{E}_0\) are induced with lightweight geometric rules on the segmentation masks.
Proximity (\emph{near}) thresholds the normalized center-distance between objects, containment (\emph{in}) checks whether one mask substantially encloses another, and support (\emph{on}) uses consistent vertical ordering and boundary contact in image space.
\\[4pt]
\textbf{Online semantic-graph state update.}
At each timestep \(t>0\), we track task-relevant objects from \(\mathcal{G}_{t-1}\) to the new observation \(o_t\) using Cutie~\cite{cutie_cvpr2024}, which updates the per-view mask groundings of existing nodes.
To recover from tracker drift and to discover new objects, we additionally run the instance segmenter on \(o_t\) and pass candidate masks through the same relevance filtering and multi-view association procedure as described in semantic-graph state initialization phase.
When a candidate region corresponds to an existing node, we merge it as a correction; otherwise, we add it as a new node only if it is deemed task-relevant and then refresh edges using relation induction rules.
\vspace{0.05in}
\subsection{Code-as-Planner}
\label{sec:codegraph_planner}

\noindent\textbf{Constructing programmatic planner.}
We represent the planner as an executable Python program \(\mathcal{P}\) that is repeatedly invoked online during manipulation deployment.
Given the current semantic-graph, it outputs the next subtask instruction and the subtask-relevant objects that should be grounded for action reasoning:
\begin{equation}
\big(l_t^{\mathrm{sub}}, \mathcal{O}_t^{\mathrm{rel}}\big)=\mathcal{P}(\mathcal{G}_t).
\end{equation}
To instantiate \(\mathcal{P}\) for a long-horizon instruction \(l\), we follow programmatic embodied control~\cite{codeaspolicies_icra2023,inner_monologue_corl2022,progprompt_icra2023} and prompt an LLM once at task initialization with \(l\), the graph schema, and the initial graph \(\mathcal{G}_0\).
The LLM synthesizes a task-specific Python planner with three components.
First, it generates helper routines that implement common graph queries (e.g., filtering objects by class and attributes, locating the container an object is currently \emph{in}, checking whether the robot is \emph{holding} an object, and selecting an empty buffer container).
Second, it defines lightweight boolean predicates that encode task constraints and progress conditions over \(\mathcal{G}_t\) (e.g., single-object holding, exclusivity constraints for containers, and relation checks for \emph{in}/\emph{on}/\emph{holding}).
Third, it implements a main \texttt{policy(graph)} function that uses these primitives to select the next subtask and returns \(\big(l_t^{\mathrm{sub}}, \mathcal{O}_t^{\mathrm{rel}}\big)\).

We instruct the LLM to first identify task roles and ordering constraints from \(\mathcal{G}_0\), and then implement a reactive program that grounds roles from semantic fields (class names, attributes, relations).
Crucially, the code maintains a lightweight persistent list \texttt{graph.task\_memory} that records which subtasks have already been completed, preventing redundant execution across repeated replanning calls.
This one-time synthesis amortizes LLM/VLM inference cost and yields a planner that can be executed cheaply without keeping a large LLM/VLM in the loop.
\\[6pt]
\textbf{Planner structure and task memory maintenance.}
During deployment, \(\mathcal{P}\) is called after each executed action chunk with the latest graph \(\mathcal{G}_t\).
At each call, the synthesized \texttt{policy(graph)} queries \(\mathcal{G}_t\) and uses simple graph predicates (e.g., \emph{holding} and \emph{in}) to check whether the current subtask has already been satisfied.
If so, it appends the subtask to \texttt{graph.task\_memory} and advances to the next unfinished subtask; otherwise, it re-issues the same instruction.
It then returns \(\big(l_t^{\mathrm{sub}}, \mathcal{O}_t^{\mathrm{rel}}\big)\), where \(\mathcal{O}_t^{\mathrm{rel}}\) contains only the objects needed to ground the immediate decision.
\\[6pt]
\textbf{Example (swap cups).}
Consider swapping two cups, starting with the black cup (\Cref{fig:architecture}(b)).
The synthesized \texttt{policy(graph)} performs one-time role binding, identifying \texttt{black\_cup}, \texttt{other\_cup}, their start plates, and an empty \texttt{buffer\_plate} for temporary storage, and caches a three-step \texttt{swap\_plan} (respecting container exclusivity) in \texttt{graph.task\_memory}.
At runtime, it scans \texttt{swap\_plan} and emits the first unfinished transfer, choosing between \emph{pick up} and \emph{put \dots\ inside \dots} based on \emph{holding}/\emph{in} predicates and returning only the subtask-relevant objects for grounding.
We use \(\mathcal{O}_t^{\mathrm{rel}}\) to construct the clutter-free visual--textual cues for the VLA policy described next.
Note that the planner only reads \(\mathcal{G}_t\); graph construction and online updates are handled by \Cref{sec:semantic_graph}.

Overall, we alternate between updating \(\mathcal{G}_t\), invoking \(\mathcal{P}\), and executing one action chunk with the VLA, which keeps planning reactive while avoiding VLM-in-the-loop inference.
Expressing planning as executable code over \(\mathcal{G}_t\) enables efficient and interpretable long-horizon subtask reasoning, while the explicit \(\mathcal{O}_t^{\mathrm{rel}}\) provides grounded context that improves robustness under clutter.

\begin{figure*}[t]
    \centering
    \includegraphics[width=0.7\linewidth]{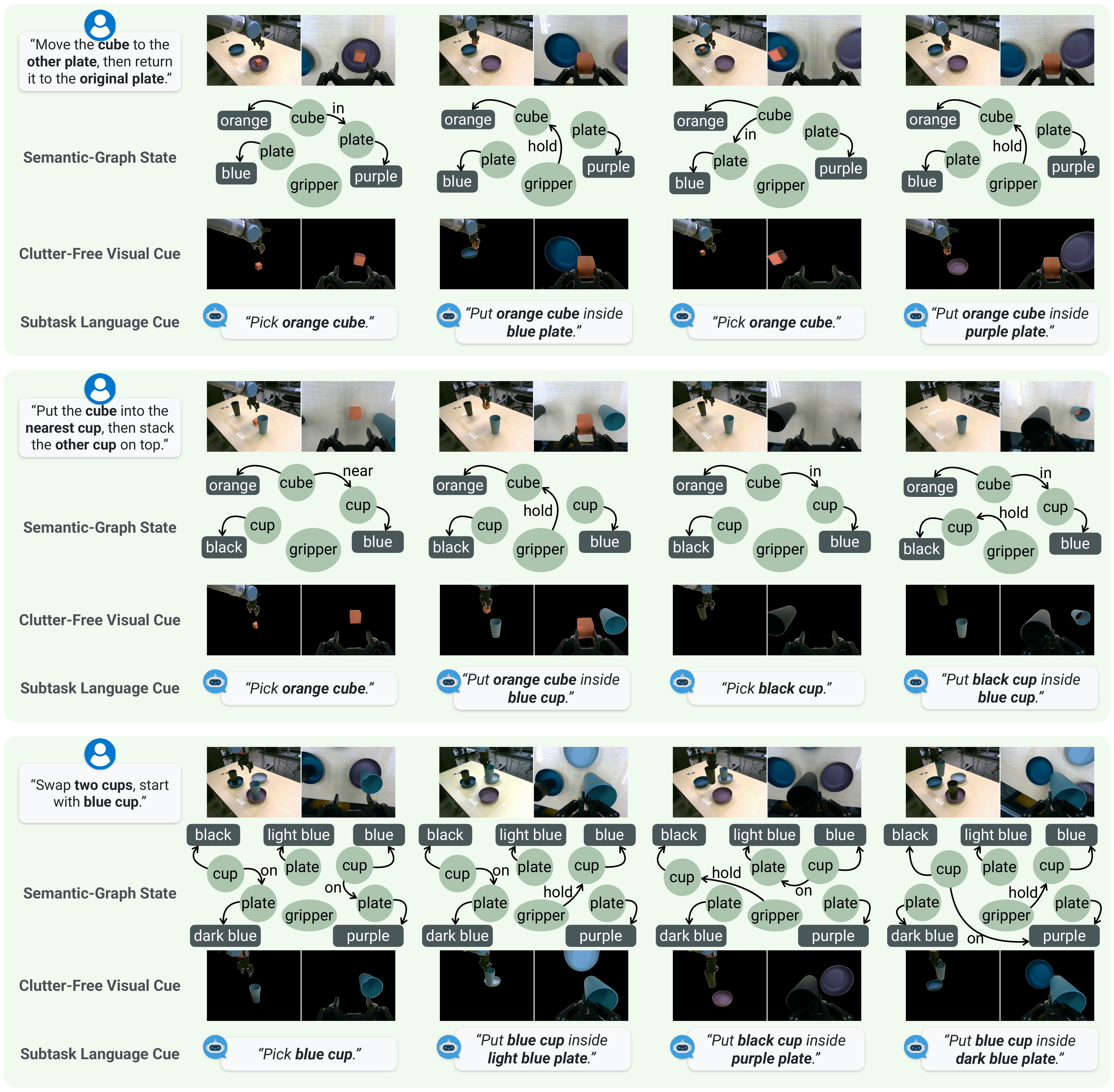}
    \vspace{-0.6em}
    \caption{\textbf{Qualitative rollouts of \model on our three real-world tasks (Pick-and-Place Twice, Place-and-Stack, and Swap Cups).} For each task, we show multi-view RGB inputs with the overall instruction, the semantic-graph state \(\mathcal{G}_t\), and the progress-guided prompts used by the VLA policy: clutter-free visual cues that retain only subtask-relevant objects and the planner-produced subtask language cues.}
    \label{fig:quali_res}
    \vspace{-0.15in}
\end{figure*}

\subsection{Clutter-Free Visual-Language Prompting}
\label{sec:vl_prompt}

Long-horizon tasks are often executed in cluttered workspaces, where many visible objects are irrelevant to the current decision and can distract a VLA from correct grounding.
We therefore convert the planner output \(\big(l_t^{\mathrm{sub}}, \mathcal{O}_t^{\mathrm{rel}}\big)\) together with the semantic-graph state \(\mathcal{G}_t\) into a progress-guided visual--language prompt for the VLA executor.
\\[4pt]
\textbf{Subtask language cue.}
Following hierarchical long-horizon systems that guide low-level execution with subtask-level language~\cite{hirobot_icml2025,mindexplore_iccv2025}, we condition the VLA executor on the planner-produced instruction \(l_t^{\mathrm{sub}}\) instead of the long-horizon task description \(l\).
Keeping \(l_t^{\mathrm{sub}}\) short and imperative focuses the language context on the immediate decision and aligns it with the subtask-relevant object set \(\mathcal{O}_t^{\mathrm{rel}}\), improving action reasoning in cluttered scenes.
\\[4pt]
\textbf{Clutter-free visual cue.}
Subtask-level language reduces linguistic ambiguity, but the VLA executor still has to perform fine-grained visual grounding from raw RGB observations, which has been shown to be brittle under clutter and distractors~\cite{byovla_icra2025}.
We address this brittleness by constructing a clutter-free observation that retains only the subtask-relevant objects \(\mathcal{O}_t^{\mathrm{rel}}\).
For each view \(v\), let \(m_{i,t}^v\) denote the binary segmentation mask of object \(i\) rendered in that view by \(\mathcal{G}_t\).
We define the view-specific retention mask as:
\begin{equation}
M_t^v = \max_{i\in \mathcal{O}_t^{\mathrm{rel}}} m_{i,t}^v,
\end{equation}
which keeps only the subtask-relevant objects.
We then form a clutter-free image by blacking out all other pixels:
\begin{equation}
\tilde{I}_t^v = I_t^v \odot M_t^v,
\end{equation}
and define the clutter-free multi-view observation as \(\tilde{o}_t=[\tilde{I}_t^1,\dots,\tilde{I}_t^n]\).
\\[4pt]
\textbf{VLA execution.}
The executor VLA predicts the next action chunk from \((\tilde{o}_t, s_t, l_t^{\mathrm{sub}})\), and we apply the same prompting during fine-tuning to match the training and deployment inputs.
By exposing only subtask-critical visual evidence and suppressing distractors, the prompting improves clutter-robust action reasoning of downstream VLA policies.
\\[4pt]
\textbf{Training objective.}
We fine-tune the VLA via imitation learning using training inputs that match the progress-guided executor inputs used at test time.
For each teleoperated trajectory paired with a long-horizon instruction \(l\), we first build the semantic-graph \(\mathcal{G}_t\) from the recorded observations using our semantic-graph construction and online update pipeline.
We then execute the synthesized code planner \(\mathcal{P}(\mathcal{G}_t)\) at each timestep to produce \(\big(l_t^{\mathrm{sub}}, \mathcal{O}_t^{\mathrm{rel}}\big)\), and mask the multi-view images to form the clutter-free observation \(\tilde{o}_t\).
This converts each long-horizon demonstration into a set of subtask-conditioned, object-focused training tuples.
We optimize the VLA parameters \(\theta\) on \(\mathcal{D}=\{(\tilde{o}_i, s_i, \tau_i, l_i^{\mathrm{sub}})\}_{i=1}^{N}\) via maximum-likelihood estimation:
\begin{equation}
\max_\theta \mathbb{E}_{(\tilde{o}, s, \tau, l^{\mathrm{sub}})\sim \mathcal{D}}\left[\log \pi_\theta (\tau|\tilde{o},s,l^{\mathrm{sub}})\right].
\end{equation}

%% file: sections/04_experiments.tex
To comprehensively evaluate \model, we design our experiments to answer three questions:\\
\textbf{Q1.} Can \model complete non-Markovian long-horizon tasks more reliably than strong VLA baselines, especially in cluttered scenes?\\
\textbf{Q2.} Does programmatic planning over a persistent semantic-graph enable more efficient progress estimation than history-enabled VLAs or VLM-in-the-loop hierarchies?\\
\textbf{Q3.} How much does progress-guided visual--textual prompting improve clutter-robust action reasoning during execution?

\subsection{Experimental Setups}

\textbf{Tasks.}
We evaluate \model on a suite of three real-world tabletop manipulation tasks that require reasoning over past observations and clutter-robust action reasoning, inspired by non-Markovian long-horizon benchmarks~\cite{hamlet_iclr2026}:\\[4pt]
\textbf{(i) Pick-and-Place Twice.}
Two plates are placed on the tabletop, and a cube is initialized on a randomly chosen plate.
The robot must pick the cube and place it onto the other plate, then pick it again and return it to the original plate.
This task is history-dependent because both the initial and terminal configurations are visually similar, so a policy can prematurely predict that it is already done.\\[4pt]
\textbf{(ii) Place-and-Stack.}
Two cups are placed on opposite sides of the workspace, and a cube is initialized near one of the cups.
The robot must first place the cube into the nearest cup, then pick up the other cup and stack it onto the cup that contains the cube.
After the cube is inserted, it is completely occluded, so selecting the empty cup to stack depends on past observations rather than the latest RGB input alone.\\[4pt]
\textbf{(iii) Swap Cups.}
Three plates are placed on the tabletop, and two cups (black and blue) are randomly placed on two of them, leaving one plate empty as a temporary buffer.
The goal is to swap the positions of the two cups, with the additional constraint that the policy must start by moving either the black cup or the blue cup (two task variants).
This requires memorizing the initial cup--plate assignments and executing multi-step swaps without reverting completed steps.

\begin{table*}[t]
    \centering
    \caption{Comparison of methods on real-world non-Markovian long-horizon tasks. All reported values are success rates (\%).}
    \label{tab:main_results}
    \small
    \setlength{\tabcolsep}{4pt}
    \renewcommand{\arraystretch}{1.05}
    \begin{tabular}{l | cc | cc | cc | c}
        \toprule
        \multirow{2}{*}{Method} & \multicolumn{2}{c|}{Pick-and-Place Twice} & \multicolumn{2}{c|}{Place-and-Stack} & \multicolumn{2}{c|}{Swap Cups} & \multirow{2}{*}{Avg.} \\
        \cmidrule(lr){2-3}\cmidrule(lr){4-5}\cmidrule(lr){6-7}
        & PnP Once & Success Rate & Drop Cube & Success Rate & Stage Cup & Success Rate & \\
        \midrule
        $\pi_0$ FAST~\cite{fast_2025} & 0 & 0 & 0 & 0 & 0 & 0 & 0.0 \\
        $\pi_0$~\cite{pi0_2024} & 0 & 0 & 60 & 40 & 55 & 50 & 30.0 \\
        $\pi_{0.5}$~\cite{pi05_2025} & 5 & 0 & 35 & 5 & 25 & 10 & 5.0 \\
        Gr00T N1.5~\cite{gr00t_2025} & 50 & 35 & 40 & 40 & 70 & 20 & 31.7 \\
        Gr00T N1.5 + Multi-frame & 100 & 75 & 50 & 50 & 90 & 45 & 56.7 \\
        \midrule
        \textbf{\model (ours)} & \textbf{100} & \textbf{80} & \textbf{95} & \textbf{80} & \textbf{100} & \textbf{85} & \textbf{81.7} \\
        \bottomrule
    \end{tabular}
\end{table*}

\begin{figure}[t]
    \centering
    \includegraphics[width=0.6\linewidth]{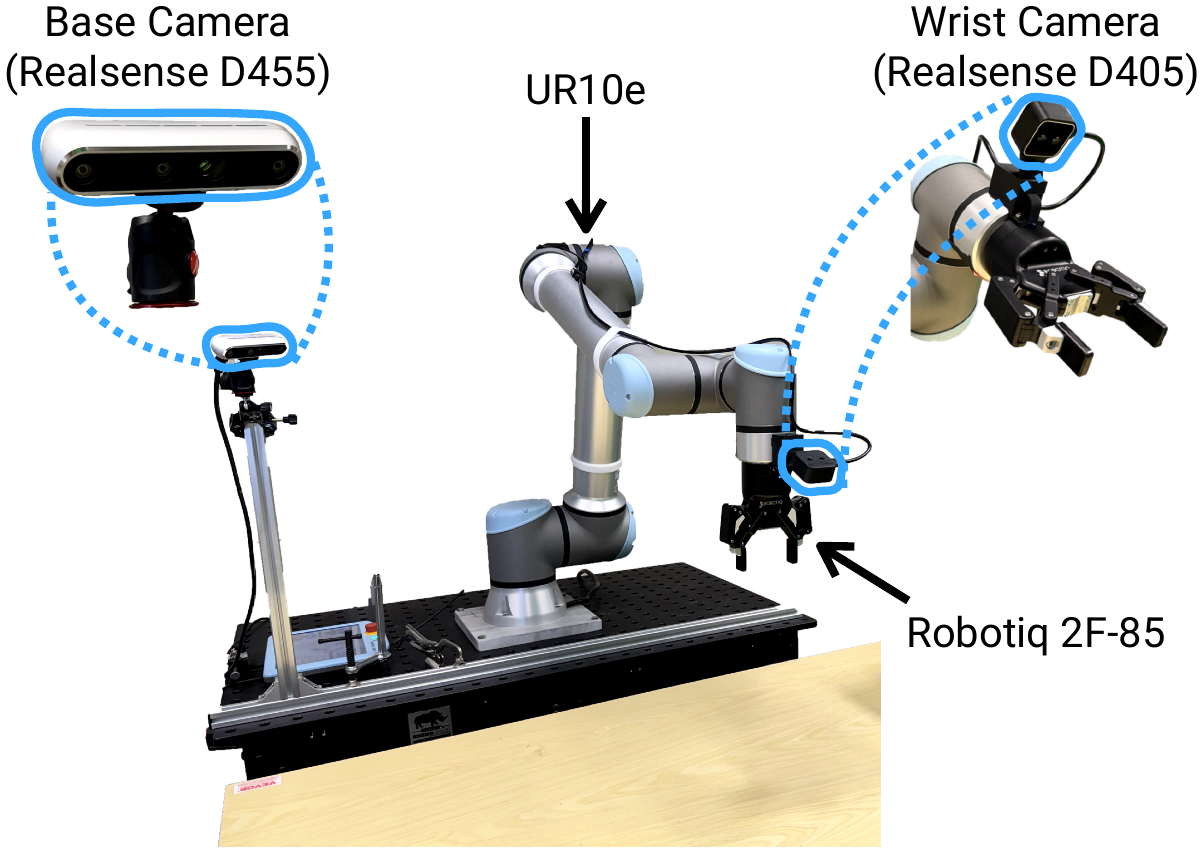}
    \vspace{-0.6em}
    \caption{Robot experimental setup on a UR10e manipulator with a parallel gripper. The system receives both a global scene observation and a wrist-view observation near the end-effector.}
    \vspace{-0.2in}
    \label{fig:robot_setup}
\end{figure}

\noindent\textbf{Demonstrations.}
We collect teleoperated demonstrations for fine-tuning all VLA models: 100 for Pick-and-Place Twice, 100 for Place-and-Stack, and 200 for Swap Cups, with 100 episodes that start with the blue cup and 100 that start with the black cup.
\\[4pt]
\textbf{Baselines.}
We compare \model against strong open VLA baselines, including $\pi_0$~\cite{pi0_2024}, $\pi_0$ FAST~\cite{fast_2025}, $\pi_{0.5}$~\cite{pi05_2025}, and Gr00T N1.5~\cite{gr00t_2025}. These baselines operate directly on raw multi-view RGB observations without progress-guided prompting. To probe the benefit of raw temporal context, we also evaluate a history-enabled Gr00T N1.5 variant that receives four past frames sampled at 1\,s intervals. We fine-tune and evaluate all baselines under the same protocol as \model, using identical task prompts and replanning cadence.
\\[6pt]
\textbf{Implementation details.}
We instantiate the VLA executor in \model with $\pi_0$~\cite{pi0_2024} and fine-tune all VLA models (ours and baselines) from their public checkpoints for 50K iterations using low-rank adaptation~\cite{lora_iclr2022} (learning rate \(1\times 10^{-5}\), batch size 128) on four NVIDIA A6000 GPUs. At test time, inference runs on a single GPU and we execute a cut-off horizon of \(H=10\) actions per forward pass before replanning. We use GPT-5~\cite{gpt4_arxiv2024} to synthesize the executable planning program once at the beginning of each task.\\[6pt]
\textbf{Hardware setup.}
All experiments are conducted on the UR10e robotic arm with a Robotiq 2F-85 parallel-jaw gripper (\Cref{fig:robot_setup}). We capture RGB observations from two synced camera streams, including a fixed camera placed in an over-shoulder viewpoint of the robot, and a wrist-mounted camera placed at the wrist--gripper interface.
During teleoperation, we record demonstrations and log observations at 10\,Hz.

\subsection{Main Results}
\Cref{tab:main_results} reports success rates across three real-world non-Markovian tasks, including intermediate milestones that capture partial progress. \model consistently achieves the highest performance, significantly outperforming all baselines, including the strongest history-augmented VLA. Improvements in both full-task and partial completion metrics suggest that \model not only executes subtasks more reliably in clutter but also maintains more consistent long-horizon progress. Representative rollouts are shown in \Cref{fig:quali_res}.

Among the baselines, $\pi_0$ FAST fails on all tasks, suggesting that autoregressive action decoding may be brittle for long-horizon closed-loop control, where greedy predictions can collapse to inert behaviors under ambiguous observations. In contrast, VLAs with diffusion or flow-based decoders appear more stable, and conditioning on temporal context further improves performance. However, even the strongest history-augmented baseline remains substantially below \model, indicating that unstructured temporal context alone is insufficient to resolve non-Markovian dependencies or clutter-induced grounding errors without explicit progress reasoning and structured visual cues.

\subsection{Ablation Analysis}
\label{sec:exp:ablation}
Unless otherwise stated, we conduct all ablation studies on the Swap Cups task, which requires both non-Markovian reasoning and clutter-robust execution.
\\[4pt]
\textbf{Efficiency benefits of Code-as-Planner.}
To evaluate the efficiency and reliability of Code-as-Planner, we compare it with two VLM-based planners that infer the next subtask online at each replanning step. Both use the latest multi-view RGB observation, while one additionally receives our semantic-graph progress state. As shown in \Cref{tab:ablation_planner}, the progress state improves VLM planning, but RGB-only reasoning remains insufficient in non-Markovian settings. Even so, VLM-in-the-loop planning remains slower and less reliable, whereas Code-as-Planner achieves higher success with substantially lower latency through one-time synthesis and lightweight graph queries.
\begin{table}[t]
    \centering
    \caption{Ablation on subtask planning efficiency on Swap Cups.}
    \label{tab:ablation_planner}
    \small
    \setlength{\tabcolsep}{4pt}
   \renewcommand{\arraystretch}{1.05}
\begin{tabular}{cc|cc}
\toprule
Code-as- &  Semantic  & Success & Latency \\
Planner & Graph & Rate & (sec/step) \\ \midrule
 \xmark & \xmark  & 25 & 2.967 \\
 \xmark &  \checkmark & 55 & 3.142 \\
  \checkmark & \checkmark  & 85 & 0.328 \\ 
  \bottomrule
\end{tabular}
\end{table}

\begin{table}[]
    \centering
    \caption{Ablation on Clutter-Free Visual--Language Prompting on Swap Cups.}
    \label{tab:ablation_prompt}
    \vspace{-0.6em}
    \small
    \setlength{\tabcolsep}{4pt}
    \renewcommand{\arraystretch}{1.05}
    \begin{tabular}{l| c}
        \toprule
        Method & Success rate \\
        \midrule
        \model w/o clutter-free visual & 40 \\
        \model & 85 \\
        \bottomrule
    \end{tabular}
    \vspace{-0.15in}
\end{table}
\vspace{0.1in}
\noindent\textbf{How effective is visual--textual prompting for VLA execution?}
We isolate the effect of the clutter-free visual cue by removing the subtask-relevant object set while keeping the subtask language cue unchanged.
Concretely, we train and deploy the VLA policy on the original multi-view RGB observations instead of masking to the subtask-relevant regions.
As shown in \Cref{tab:ablation_prompt}, this reduces success from 85\% to 40\%, indicating that progress-guided perceptual inputs are important for clutter-robust action reasoning.

%% file: sections/05_conclusion.tex
We introduced \model, a framework that enables long-horizon manipulation by combining structured scene memory with programmatic planning and grounded visual–language prompting. Experiments on real-world tasks show that this design improves task success while reducing planning latency compared to VLM-in-the-loop and memory-augmented VLA baselines.
\\[4pt]
\noindent\textbf{Limitations.}
Our pipeline relies on foundation models for both semantic-graph construction and program synthesis, so performance depends on their capabilities and failure modes.
In particular, the attributes and relations inferred by the VLM can be sensitive to camera viewpoints and visual quality, and the generated Code-as-Planner requires careful prompt design to ensure reliable, executable logic.
Future work includes developing an open-world semantic-graph generation module that streamlines the construction of persistent semantic-graph state, and automated verification and self-checks for synthesized code-based planners.